\documentclass[10pt,twocolumn,letterpaper]{article}
\usepackage[pagenumbers]{cvpr} 
\usepackage{booktabs}
\usepackage{multirow}
\usepackage[accsupp]{axessibility}
\usepackage[dvipsnames]{xcolor}
\usepackage{dcolumn}
\newcolumntype{d}[1]{D{.}{.}{#1}}
\makeatletter
\newcolumntype{B}[3]{>{\boldmath\DC@{#1}{#2}{#3}}c<{\DC@end}}
\makeatother
\newcommand\mc[1]{\multicolumn{1}{c}{#1}}
\newcommand\boldc[1]{\multicolumn{1}{B{.}{.}{2.4}}{#1}}
\newcommand{\spm}[1]{{\scriptscriptstyle\pm#1}}
\newcolumntype{C}[1]{>{\centering\arraybackslash}m{#1}}

\definecolor{cvprblue}{rgb}{0.21,0.49,0.74}
\usepackage[pagebackref,breaklinks,colorlinks,citecolor=cvprblue]{hyperref}
\usepackage{listings}

\def\paperID{15960} % *** Enter the Paper ID here
\def\confName{CVPR}
\def\confYear{2024}

\title{How to Train Neural Field Representations: A Comprehensive Study \\ and Benchmark}

\author{Samuele Papa\textsuperscript{1,2}, Riccardo Valperga\textsuperscript{1}, David Knigge\textsuperscript{1,2}, Miltiadis Kofinas\textsuperscript{1}, Phillip Lippe\textsuperscript{1}, \\ Jan-Jakob Sonke\textsuperscript{2}, Efstratios Gavves\textsuperscript{1} \\
\textsuperscript{1} University of Amsterdam, 
\textsuperscript{2} Netherlands Cancer Institute\\
{\tt\small s.papa@\{uva,nki\}.nl}, {\tt\small \{r.valperga, dm.knigge, m.kofinas, p.lippe\}@uva.nl}, \\ {\tt\small j.sonke@nki.nl}, {\tt\small e.gavves@uva.nl} 
}

\begin{document}
\maketitle
\begin{abstract}
Neural fields (NeFs) have recently emerged as a versatile method for modeling signals of various modalities, including images, shapes, and scenes.
Subsequently, a number of works have explored the use of NeFs as representations for downstream tasks, e.g. classifying an image based on the parameters of a NeF that has been fit to it.
However, the impact of the NeF hyperparameters on their quality as downstream representation is scarcely understood and remains largely unexplored.
This is in part caused by the large amount of time required to fit datasets of neural fields.

In this work, we propose a JAX-based library\footnote{\href{https://fit-a-nef.github.io/}{https://fit-a-nef.github.io/}} that leverages parallelization to
enable fast optimization of large-scale NeF datasets, resulting in a significant speed-up.
With this library, we perform a comprehensive study that investigates
the effects of different hyperparameters on fitting NeFs for
downstream tasks. In particular, we explore the use of a shared initialization, the effects of \textit{overtraining}, and the expressiveness of the network architectures used.
Our study provides valuable insights on how to train NeFs and offers guidance for optimizing their effectiveness in downstream applications. 
Finally, based on the proposed library and our analysis, we propose Neural Field Arena, a benchmark consisting of neural field variants of popular vision
datasets, including MNIST, CIFAR, variants of ImageNet, and ShapeNetv2.
Our library and the Neural Field Arena will be open-sourced to introduce
standardized benchmarking and promote further research on neural fields.
\end{abstract}

\begin{figure*}[h]
\begin{subfigure}[b]{\textwidth}
         \centering
         \includegraphics[width=\linewidth]{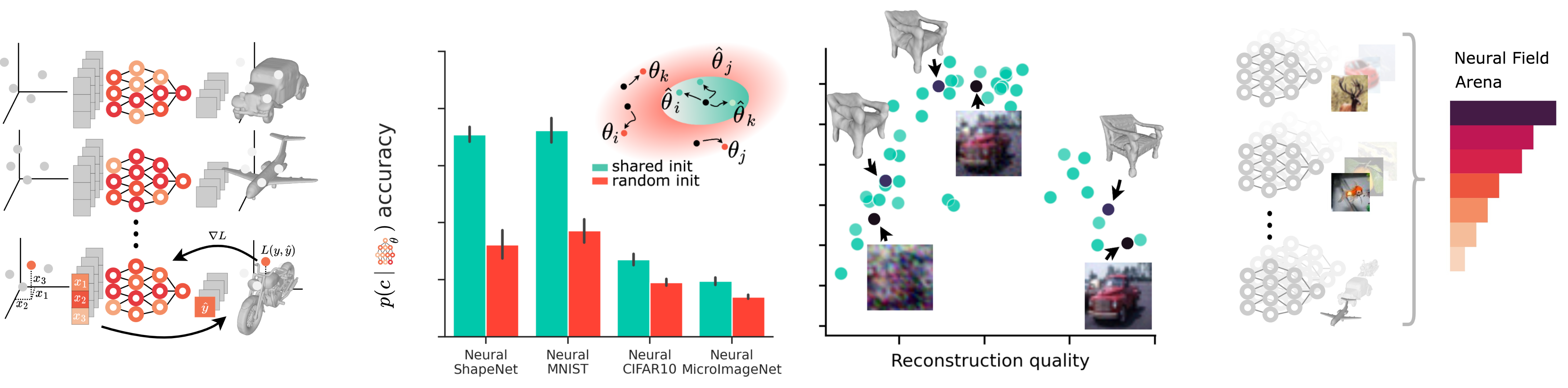}
     \end{subfigure}
    \centering
         \begin{subfigure}[b]{0.27\textwidth}
         \centering
         \caption{}
         \label{fig:fig1a-fitting}
     \end{subfigure}
     \hspace{1mm}
     \begin{subfigure}[b]{0.45\textwidth}
         \centering
         \caption{}
         \label{fig:fig1b-studies}
     \end{subfigure}
     \hspace{2mm}
     \begin{subfigure}[b]{0.23\textwidth}
         \centering
         \caption{}
         \label{fig:fig1c-benchmark}
     \end{subfigure}
    \caption{To investigate the use of Neural Fields (NeFs) as representations, we propose \texttt{Fit-a-NeF} (\ref{fig:fig1a-fitting}), a JAX-based library for efficient parallelized fitting of NeFs to datasets of signals, obtaining a $100{-}1,300\times$ speed-up. Owing to this efficiency, we are able to uncover the impact of NeF hyperparameter choices on their usability as representations -- evaluated in downstream classification (Fig. \ref{fig:fig1b-studies}) -- obtaining two important insights. First, \ref{fig:fig1b-studies}-left, it is vital to \textit{group NeFs in parameter-space}, which we propose to enforce by sharing their network initializations. Second, \ref{fig:fig1b-studies}-right, \textit{improved reconstruction quality does not necessarily result in improved representation quality}, implying an optimal combination of NeF expressivity and optimization for learning on NeFs. Incorporating these insights, we create a suite of NeF-based variants of classical CV datasets, Fig. \ref{fig:fig1c-benchmark}. We bundle these \textit{Neural Datasets} into a benchmark for learning on Neural Fields -- which we name Neural Field Arena -- hoping to enable standardized comparison in order to promote further research into this field.}
    \label{fig:1}
\end{figure*}

%%%%%%%%%%%%%%%%%%%%%%%%%%%%%%%
\section{Introduction}
\label{sec:intro}
%%%%%%%%%%%%%%%%%%%%%%%%%%%%%%%

Neural Fields (NeFs) \cite{xie2022neural} have recently emerged as a versatile method for reconstructing discretely sampled continuous signals, with great success on various modalities from images~\citep{stanley2007compositional, sitzmann2020implicit, tancik2020fourier}, videos~\citep{park2021nerfies}, 3D scenes and shapes~\citep{mescheder2019occupancy, park2019deepsdf, mildenhall2021nerf, yu2021pixelnerf}, to medical imaging~\citep{zha2022naf, papa2023neural}, audio~\citep{sitzmann2020implicit, gao2021objectfolder}, and physical systems \cite{raissi2019physics,yin2022continuous,kofinas2023latent}.
The continuous nature of coordinate-based models $f_\theta\colon x \mapsto y, x \in \mathbb{R}^{d}, y \in \mathbb{R}^c$ (e.g. $d=2$ for images and $d=3$ for 3D shapes), allows for modeling signals without the limitations in resolution inherent in discrete representations; their capacity for modeling fine detail is only limited by the expressivity of the underlying NeF architecture. 
Furthermore, some modalities, such as scenes and shapes, do not have array representations that can be easily adapted to existing deep learning methods. 
Doubling down on their flexibility, a growing body of literature~\cite{dupont2022data, bauer2023spatial, navon2023equivariant, zhang2023neural, erkocc2023hyperdiffusion} has further proposed to use NeF parameterizations $f_\theta$ directly as \textit{representations of data}. They are used to model images, videos, or shapes not as tensors but as functions on which to run downstream tasks (\emph{e.g.}, classify the function of an image instead of the pixel tensor of an image). In this setting, the parameters $\theta$ of a NeF --sometimes along with their computational structure-- serve as input for a downstream model.

Despite their promise, NeFs lag behind as image representations, while showing promise for shape datasets~\cite{navon2023equivariant, zhang2023neural}.
We argue that two challenges contribute to their lack of performance: a practical one, and a theoretical one.
The practical challenge is that a NeF is a neural network fit on a single discretely sampled signal, for instance, the pixels of an image.
Fitting a new NeF for every sample in a dataset means training a new neural network from scratch, thus scaling up to whole datasets --especially large datasets with high-resolution signals-- is prohibitively time-consuming \cite{dupont2022data,navon2023equivariant, erkocc2023hyperdiffusion}.
The theoretical challenge stems from the fact that in this setting we treat the parameters of a neural network as a signal representation, and given our lack of understanding of how neural networks behave as a function of their architecture and optimization, we do not know what makes the parameters of a NeF a good representation for downstream tasks.
Moreover, the scaling inefficiencies described above hinder us from forming hypotheses and testing them experimentally, inhibiting systematic research on NeF representations.

In this work, we address both challenges. 
Firstly, the existing libraries for learning neural networks are not necessarily optimal for fitting NeFs. To this end, 
we propose a JAX-based library \cite{jax2018github, flax2020github} --which we dub \texttt{Fit-a-NeF}-- that leverages both in-device and across-device parallelization to reduce the time required for NeF fitting drastically.

Secondly, we explore the architectural considerations and optimization choices for making a NeF perform well as a representation. Equipped with an efficient library, we form structured hypotheses on the nature of parameter space --i.e. the space of the parameters of neural networks-- and test them empirically.

In this work, we posit and experimentally verify that by constraining distinct NeFs to be close in parameter space, we improve the ability of downstream models to learn on the resulting sets of parameters.
Further, we posit and experimentally verify that reconstruction quality with NeFs is not equivalent to the quality of NeF representations. From our analysis, we note that generalization to off-grid sampling should be used as an additional measure to gauge the quality of NeF representations.

Following our previous argument, we fit NeFs on popular datasets, which we call \emph{Neural Datasets}, starting with MNIST, CIFAR, a modified version of ImageNet, and ShapeNet.
To foster further research on NeF representations, we set up \emph{Neural Field Arena}, a series of benchmarks on these Neural Datasets analogous to the benchmarks for the original datasets. In short, our contributions are as follows.\looseness=-1
\begin{itemize}
    \item We provide a JAX-based library to quickly and easily fit millions of neural fields for different modalities, including images and shapes, using accelerators with parallelization on and across several devices. 
    \item We use the library to perform a thorough investigation on how to obtain NeFs for datasets of signals to facilitate better downstream performance, expressed in downstream classification accuracy. We motivate and subsequently show that limiting parameter-space distances between NeFs during optimization is critical to obtaining NeFs that may serve as viable downstream representations. Furthermore, we demonstrate that the design choices for learning strong NeF representations are not necessarily the same as the design choices for good NeF reconstructions.
    \item We process NeF variants of popular computer vision datasets, creating a benchmark for Neural Field-based learning that we coin \emph{Neural Field Arena} in the hopes of streamlining research in this direction.
\end{itemize}
We open-source code for the experiments and provide Neural Field Arena, a benchmark for learning on Neural Fields.

%%%%%%%%%%%%%%%%%%%%%%%%%%%%%%%%%%%%%%%%%%%%%%%%%%%%%%%%%%%
\section{Representations in Parameter Space}
%%%%%%%%%%%%%%%%%%%%%%%%%%%%%%%%%%%%%%%%%%%%%%%%%%%%%%%%%%%

Neural Fields are functions $f_\theta\colon x \mapsto y$, parameterized by a neural network whose parameters $\theta$ we optimize to reconstruct the continuous signal $y$ on coordinates $x$.
As with regular neural networks, fitting NeFs relies on gradient descent minimization.
NeF representations are simply the parameters $\theta$ -- possibly alongside their corresponding computational graph -- that result from this optimization.
As a consequence of the above, NeF representations inherit the intricacies and complexities of the topological space of neural networks, namely the extremely non-convex and high-dimensional optimization landscapes resulting from severe overparameterization. Any downstream model that uses NeFs as signal representation is then required to learn on this highly complex space of representations.

To be more specific, for any two functions $f_1$ and $f_2$ (which we would like to fit with NeFs) that are $\epsilon-$close in function space, it is not always theoretically possible to fit neural networks to these respective functions that are also close in their parameter space~\cite{petersen1806topological}.
In other words, two signals that are similar might not have NeF representations which are themselves similar.
Furthermore, the same neural network can be realized with parameters that are far apart in parameter space \cite{hecht1990algebraic, brea2019weight, ainsworth2022git}.
Importantly, a recent line of work shows that it is possible to align such neural networks that are close in function space but distant in parameter space, for example by finding optimal permutations in the weight space \cite{ainsworth2022git} or finding optimal transport solutions to (soft) neuron alignment \cite{singh2020model}. Together, these properties dramatically complicate the task of learning on NeFs.

%%%%%%%%%%%%%%%%%%%%%%%
\subsection{Grouping Representations in Parameter Space}
%%%%%%%%%%%%%%%%%%%%%%%

\citet{ainsworth2022git} posit that --modulo the permutation symmetries naturally present in neural networks-- the local minima reached by optimization of different randomly initialized neural network weights form a convex basin. Following these insights, \cite{zhou2023permutation, navon2023equivariant, zhang2023neural} show that downstream models that are invariant to these ambiguities improve performance dramatically over models that do not account for them. Both of these findings support the intuition that enforcing alignment in parameter space simplifies downstream learning. Our hypothesis is that the smaller pairwise distances in parameter space may simplify training because grouped NeF representations span a smaller region in parameter space, preventing, at least partially, very different NeFs from corresponding to semantically similar signals.

One possible solution is to train all NeFs for the same dataset using the same randomly selected but shared initialization instead of randomly initializing every NeF separately, see Fig. \ref{fig:fig1b-studies}.
Intuitively, with such a shared initialization, the NeF representations do not stray away too far, thus making subsequent downstream learning easier.
Based on that, we form our first research question.

\emph{RQ1: Do neural datasets obtained starting from a shared initialization lead to better downstream performance, when measured in terms of accuracy?}

%%%%%%%%%%%%%%%%%%%%%%%
\subsection{Reconstruction vs. Representation Quality}
%%%%%%%%%%%%%%%%%%%%%%%

The primary purpose of NeFs in literature has been reconstructing the target signal, whether for image representation, shape modeling, or 2D-to-3D scene reconstruction with NeRFs \cite{sitzmann2020implicit, tancik2020fourier, mildenhall2021nerf}.
As a consequence, much of the research into NeFs primarily focused on improving reconstruction quality \cite{hertz2021sape, martel2021acorn, landgraf2022pins}, or providing architectures with favorable optimization properties \cite{fathony2020multiplicative}.
However, improvements that lead NeFs to better reconstruction quality do not necessarily yield NeF parameters that serve as good representations.
The reason is that reconstruction learning is a transductive problem where the primary goal is fitting a single function as accurately as possible.
By contrast, representation learning is an inductive problem where generalization to out-of-sample data is also important.
\subsubsection{Overtraining Neural Fields}
Our hypothesis is that by overtraining we might obtain NeFs that achieve excellent reconstruction quality whose parameters, however, form poor representations. %may travel unnecessarily far from the initialization. 
Neural Fields, like neural networks in general, are susceptible to overfitting. In NeFs, this would be reflected by sets of parameters that only approximate the underlying continuous signal well on the locations for which training data is available, and diverge from the signal elsewhere in the domain. 
%Although gradient descent leads to better reconstructions, this will eventually lead to overfitting, as NeFs are often over-parametrized when compared to the input signal. 
Such NeF representations would be potentially ambiguous, leading to detrimental downstream performance. As such, we investigate the impact of overtraining on reconstruction quality \textit{outside} the training domain by comparing with a surrogate of reconstruction quality in such locations and the number of steps that the network has been trained for. Ultimately, we assess its relation with NeF representation quality. \looseness=-1 

\emph{RQ2: Can overtraining NeFs negatively affect representation quality in terms of downstream accuracy, even if it improves the reconstruction of the given signals?}

\begin{figure}[t]
\centering
\resizebox{\linewidth}{!}{%
\begin{tabular}{@{}C{0.28\linewidth}C{0.18\linewidth}C{0.18\linewidth}C{0.18\linewidth}C{0.18\linewidth}@{}}
\toprule
Neural Dataset & \multicolumn{3}{c}{Reconstruction Quality} & \\
\cmidrule{2-4}
& Low & Medium & High & Ground Truth \\
\midrule
Neural MNIST & \includegraphics[width=1.0\linewidth]{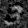} & \includegraphics[width=1.0\linewidth]{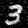} & \includegraphics[width=1.0\linewidth]{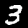} & \includegraphics[width=1.0\linewidth]{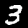} \\
Neural CIFAR10 & \includegraphics[width=1.0\linewidth]{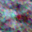} & \includegraphics[width=1.0\linewidth]{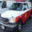} & \includegraphics[width=1.0\linewidth]{figures/figure_table/cifar10_mid.png} & \includegraphics[width=1.0\linewidth]{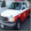} \\
Neural MicroImageNet & \includegraphics[width=1.0\linewidth]{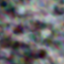} & \includegraphics[width=1.0\linewidth]{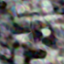} & \includegraphics[width=1.0\linewidth]{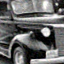} & \includegraphics[width=1.0\linewidth]{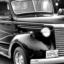}  \\
Neural ShapeNet & \includegraphics[width=1.0\linewidth]{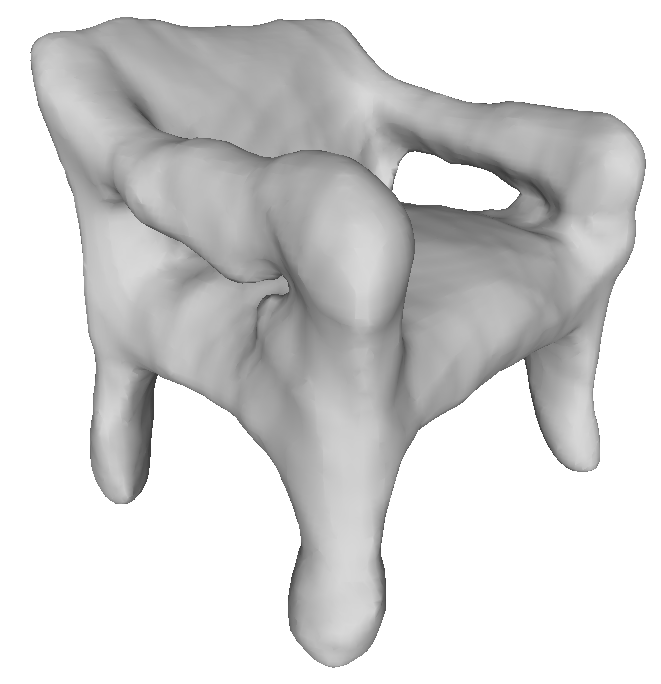} & \includegraphics[width=1.0\linewidth]{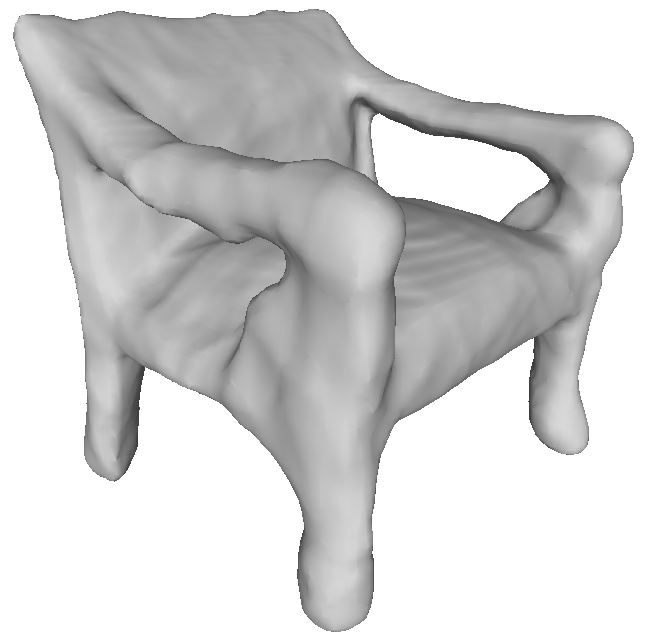} & \includegraphics[width=1.0\linewidth]{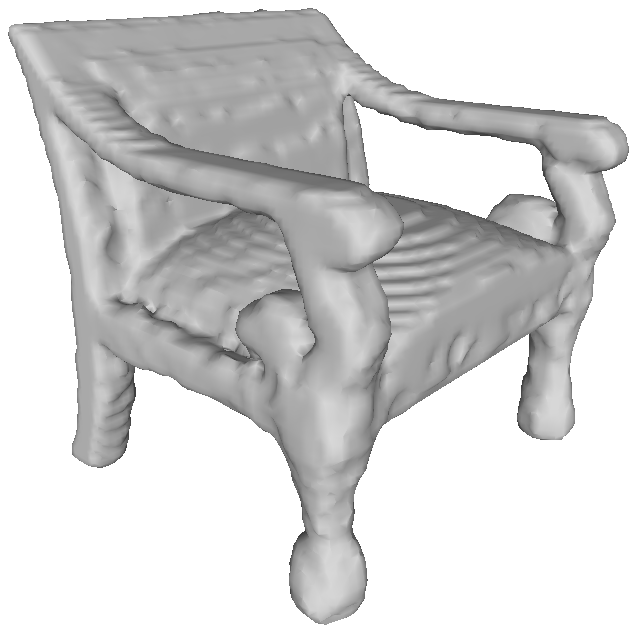} & \includegraphics[width=1.0\linewidth]{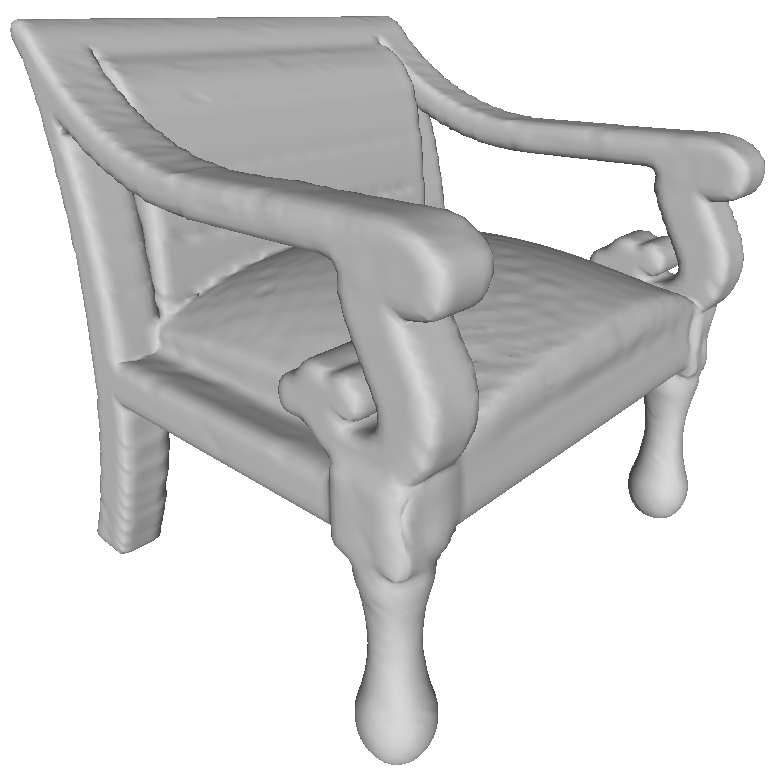} \\
\bottomrule
\end{tabular}}
\caption{From left to right, samples from neural datasets with increasing reconstruction quality. The right-most column shows the ground truth used for fitting. }
\label{table:results_example}
\end{figure}

\subsubsection{Expressivity of Neural Field Architectures}

Similar to overtraining, our hypothesis is that NeF architectures originally proposed to improve reconstruction quality might not be offering better representational benefits despite their increased ability to reconstruct signals in high fidelity. In fact, we believe overly expressive architectures might have limited use as downstream representations, since their increased expressivity intuitively transfers as increased complexity in any downstream learning.
A possible solution is to balance the expressivity of the neural network - i.e. its ability to reconstruct the signals of the dataset well - with its representation quality. As such, we propose to investigate the validity of this hypothesis on the reconstruction-representation trade-off and the viability of balancing reconstruction and representation through hyperparameter choices.

\emph{RQ3: Can the expressivity of NeF architecture be detrimental to their representation quality even if it helps with reconstruction quality?}

\section{Experimental Design}
\label{subsec:experimental_design}
%%%%%%%%%%%%%%%%%%%%%%%%%%%%%%%%%%%%%%%
We start by giving a short description of our library that we coin \texttt{Fit-a-NeF}. Then, we describe the experimental design used to test our hypotheses and answer the research questions.

\subsection{\texttt{Fit-a-NeF}: Fitting Neural Fields at Scale}
\label{sec:fit-a-nef}

The problem with fitting neural fields at scale is that training millions of small networks sequentially is not well-suited for accelerators, i.e. a GPU or a TPU, that 
are made to perform large matrix multiplications. As emphasized by \citet{dupont2022data}, fitting neural fields to large datasets can be computationally prohibitive. Modern Deep Learning frameworks offer \verb|vmap|, a higher-order function used for automatic vectorization that applies a function element-wise to arrays, enabling batch processing and efficient vectorized computations by mapping over array axes without explicit loops.
Our library uses JAX's vmap \cite{jax2018github} and Flax \cite{flax2020github} to parallelize the NeFs training process.
Together with the Just-In-Time (JIT) compilation of JAX, the library attains considerable speedups.

The primary goal of our library is to enable fast generation of Neural Datasets without the need of several GPUs. The library is sufficiently flexible to support datasets of different signals types, and new types can be integrated easily. A secondary goal is to offer a framework for loading and processing NeF representations. 

\begin{figure}
    \centering
    \includegraphics[width=.9\linewidth]{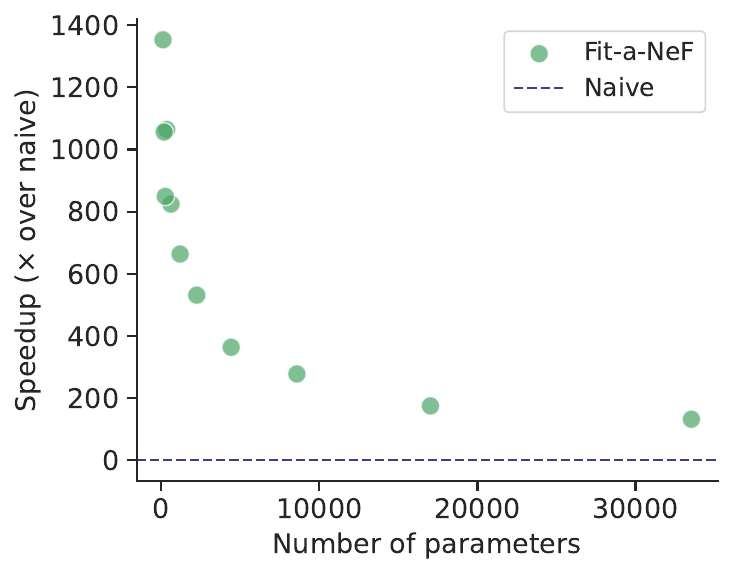}
    \caption{Speedup obtained using \texttt{Fit-a-NeF} over a naive sequential approach using SIRENs using different hidden dimensions and number of layers. The evaluation was performed using 30k samples from MNIST on an A100 GPU. Smaller networks show the biggest speedup, as the parallelization is more effective.}
    \label{fig:speedup_vs_param_count}
\end{figure}

\vspace{1mm}
\noindent\textbf{Speed comparisons.}
We show comparisons of \texttt{Fit-a-NeF} with a naive approach in Fig. \ref{fig:speedup_vs_param_count}. Specifically, we compare the time it takes to fit various neural datasets with our library and with a naive approach consisting of fitting one NeF at a time, on the same GPU. The gain depends on various factors such as the GPU memory and the size of the datapoints. For example, we can fit Neural MNIST in 17 seconds\footnote{this is true if the GPU memory allows you to train the whole MNIST in parallel, e.g., on an NVIDIA A100} vs. the 3 hours it would take with the naive approach.
In short, with \texttt{Fit-a-NeF} it took 62 GPU days on A100s to run all the experiments reported in this work, whilst a naive approach would have required approximately 30 GPU years.

\subsection{Datasets and Metrics}
In this work, we explore both shapes and images for NeF representations.
For the shape experiments, we use a 10-class version of ShapeNetv2~\cite{shapenet2015} for occupancy prediction, which we call ShapeNetv2-10. Based on the ShapeNetv2 shapes, we create the dataset by sampling 200k points at random, 100k of which are near the surface of the objects, as in~\cite{erkocc2023hyperdiffusion}.
For images, we use MNIST~\cite{lecun1998gradient} and CIFAR10~\cite{krizhevsky2009learning}. Additionally, we created a 10-class version of ImageNet~\cite{imagenet} with $64\times 64$ resolution images, which we call MicroImagenet. The latter presents a more challenging signal to fit with the neural fields.
We opt for a 10-class version of ShapeNetv2 and ImageNet for consistency. See Fig. \ref{table:results_example} for examples.

\paragraph{Classification accuracy.}
In order to address our research questions and establish the quality of NeF representations, we train a downstream model for downstream tasks.
More specifically, we choose NeF classification as a downstream task and report test accuracy, following prior work on representation learning~\cite{bengio2013representation,chen2020simple,locatello2019challenging}.
For a downstream model, it is important to use a method that respects the neural network permutation symmetries.
We choose the work of~\citet{zhang2023neural} that represents neural fields as computational graphs and processes them with standard graph networks.
We create a variant that relies on a simpler message passing neural network (MPNN).
We provide details about the classifier implementation in the appendix.

\paragraph{On-grid and off-grid reconstruction quality.} We use \textit{peak signal-to-noise ratio} (PSNR) in images and \textit{intersection over union} (IOU) in shapes to measure the quality of reconstruction or generation. Additionally, we propose to extend both to what we call their \emph{off-grid} counterparts; calculated using coordinates that are different from those used to fit the NeF to the signal.
For \textit{images}, the off-grids points lie in-between pixels, and the ground-truth image is obtained through a $\beta^{(1)}$-spline interpolation of the original image at those coordinates.
For \textit{shapes}, new points and their respective occupancy is calculated for all shapes in the dataset.

We refer to them as ``off-grid-PSNR'' and ``off-grid-IOU'', and
use PSNR and IOU only to refer to their \emph{on-grid} versions, which are computed on the coordinates used to fit the signals. In the experiments, we find the ratio between the two metrics $\frac{\texttt{off-grid-PSNR}}{\texttt{on-grid-PSNR}}$, and $\frac{\texttt{off-grid-IOU}}{\texttt{on-grid-IOU}}$ particularly informative.

\begin{table}[t]
\centering
\resizebox{\columnwidth}{!}{%
\begin{tabular}{@{}lld{2.4}d{3.2}@{}}
\toprule
Neural dataset & Initialization & \mc{Test accuracy} & \mc{Gain (\%)} \\
\midrule
\multirow{2}{*}{\textit{Neural MNISTs}} & Random & 0.37\spm{0.18} & \\
  & Shared & \boldc{0.72\spm{0.17}} & 94.69\\
\midrule
\multirow{2}{*}{\textit{Neural CIFAR10s}} & Random & 0.19\spm{0.05} & \\ 
  & Shared & \boldc{0.27\spm{0.08}} & 42.22 \\ 
\midrule
\multirow{2}{*}{\textit{Neural MicroImageNets}} & Random & 0.14\spm{0.03} & \\
  & Shared & \boldc{0.20 \spm{0.05}} & 39.93\\
\midrule
\multirow{2}{*}{\textit{Neural ShapeNets}} & Random & 0.32\spm{0.18} & \\
  & Shared & \boldc{0.71 \spm{0.10}} & 119.63 \\
\bottomrule
\end{tabular}
}
\caption{We measure the average \textit{test accuracy} (mean $\pm$ standard deviation) and percentage gain across 110 Neural Datasets trained using shared initialization and random initialization with different number of steps and hidden dimensions. }
\label{tab:shared_vs_random_init}
\end{table}

\vspace{1mm}\noindent
\textbf{Normalized Mutual Information.}
Intuitively, high-quality NeFs representations are semantically structured. We propose to use \emph{Normalized Mutual Information} (NMI) to measure the information shared between the assignments of some clustering algorithm, here \textit{k-means}, with $k$ equal to the number of classes in the dataset, and the ground truth labels of the represented signals in a Neural Dataset.

\subsection{Optimization}

To obtain NeF representations of images, we use a mean-squared error loss function of the predicted pixel intensity of a pixel coordinate compared to the ground truth pixel intensity.
For shapes, we use binary cross-entropy measuring whether a point is correctly classified to be inside or outside the shape.
We optimize the NeFs with Adam~\cite{kingma2015adam}.
The experiments consist of a vast grid search on various hyperparameters including the NeF architecture, number of steps, and learning rate.

For a given hyperparameter choice, we fit a \emph{neural dataset} to the images or shapes in the training dataset and then train a downstream classifier on it for a fixed number of epochs. The fitted Neural Dataset is, then, split into train, validation, and test subsets and the checkpoint classifier model with the highest validation accuracy is tested on the test set to obtain the test accuracy.
\subsection{Architecture}

In the main text of the paper we using SIREN~\cite{sitzmann2020implicit}. In the supplementary material, we additionally experiment with different architectures classically used in NeF literature,  RFFNets~\cite{tancik2020fourier}, and MFNs (FourierNet)~\cite{fathony2020multiplicative} to assess their impact on representation quality, as well as the effects of the learning rate and weight decay. Given a certain architecture, we also explore its depth and width.

\begin{figure}[t]
    \centering
    \includegraphics[width=\linewidth]{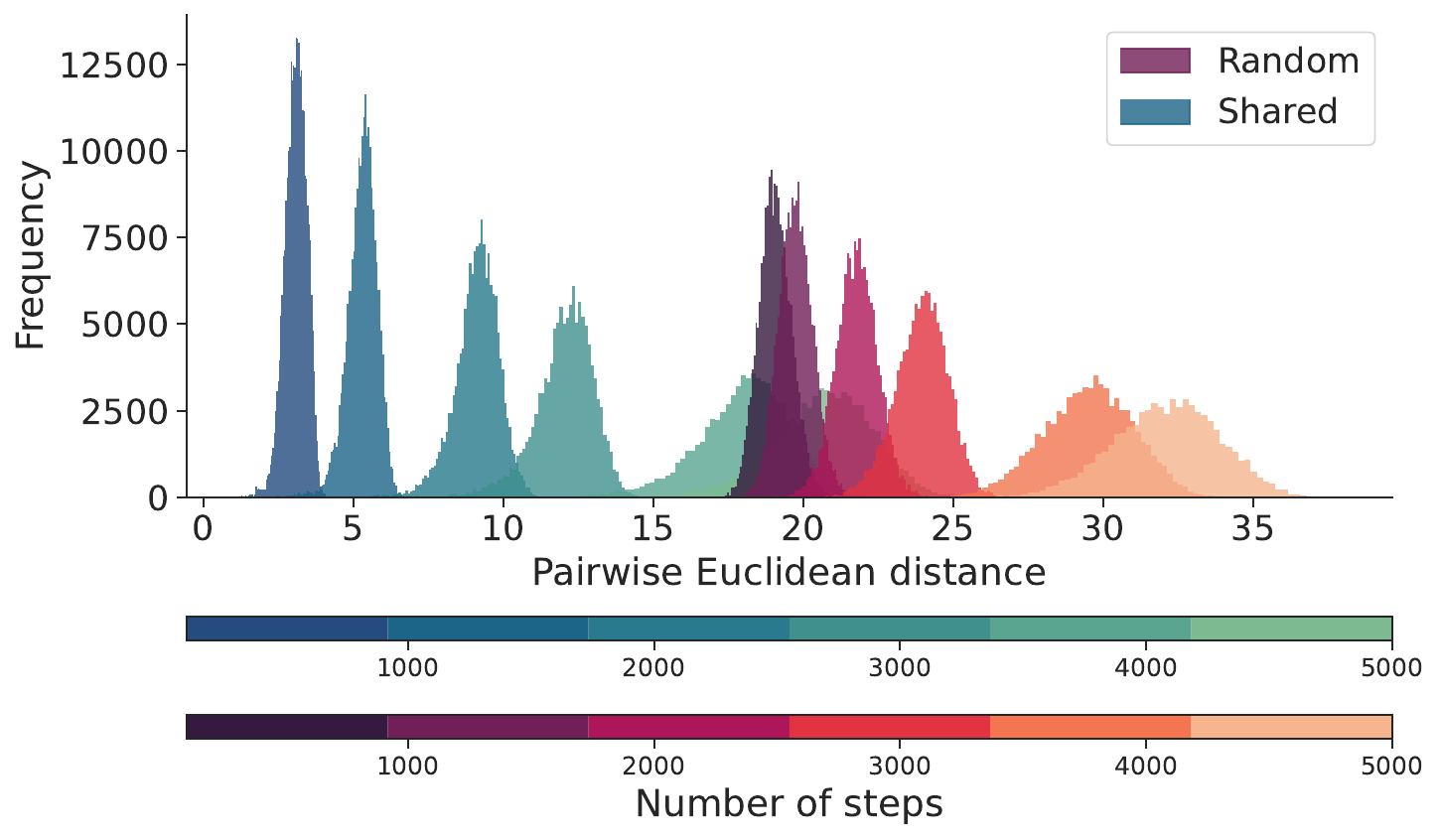}
    \caption{The histograms show the distribution of pairwise distances of the NeF representations. We fit 10 ShapeNet-10 Neural Datasets varying initialization and total number of steps. Shared initialization produces more grouped representations, and pairwise distances increase with the number of steps.}
    \label{fig:pairwise_distances}
\end{figure}

%%%%%%%%%%%%%%%%%
\section{Results and Discussion}
\label{sec:results_and_discussion}
%%%%%%%%%%%%%%%%%

Here, we describe the experiments to test the aforementioned research questions and analyze the results.
In the experiments that follow, each dot in the figures corresponds to fitting a whole dataset using a different set of hyperparameters that relates to a particular experiment.
For instance, in Fig.~\ref{fig:acc_vs_nmi_alldata}, for the ShapeNet plot we have fitted $54$ datasets for the $54$ points plotted.
We ran around $30\text{,}000$ smaller preliminary experiments to determine the importance of the hyperparameters. Following this preliminary study we ran in-depth studies with the most relevant ones.

\subsection{Shared vs. Random Initialization}

\paragraph{Description.}
We experiment with two different initialization schemes for the neural fields before fitting them to the signals.
With \emph{random} initialization we refer to the standard setting of randomly initializing the NeF afresh for every new image or shape.
With \emph{shared} initialization, we denote the approach of initially randomly sampling an initialization for the NeF parameters and subsequently using the same initialization to fit all images or shapes.

\paragraph{Analysis.}
As reported in Table \ref{tab:shared_vs_random_init}, shared initialization consistently yields higher accuracy.

To explain why shared initialization results in better performances we report in Fig. \ref{fig:pairwise_distances} the distribution of pairwise Euclidean distances between the NeF representations obtained with shared initialization and with standard initialization for different numbers of gradient descent steps.

\begin{figure}[t]
    \centering
    \includegraphics[width=\linewidth]{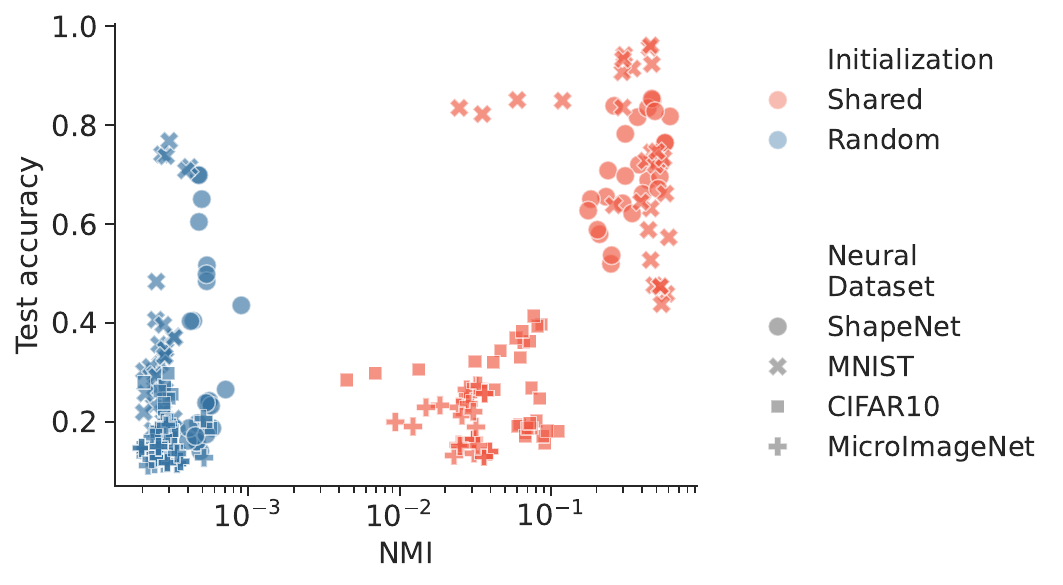}
    \caption{Results of the \textit{test accuracy} ($\uparrow$) vs \textit{NMI} ($\uparrow$) using different initialization on 220 Neural Datasets created using different hidden dimensions and the number of steps. Different datasets are stylized using different markers. Shared initialization leads to semantically structured NeF representation and, generally to better performance. The NMI of CIFAR10 and MicroImageNet are lower than those of ShapeNet and MNIST, however, are still clearly separated from their random initialization counterpart. }
    \label{fig:acc_vs_nmi_alldata}
\end{figure}

Figure \ref{fig:pairwise_distances} shows that NeF representations trained for a larger number of steps are much more spread in representation space. Given the results of our study, we posit that this spread causes a decrease in the quality of the representations, likely making separability harder.
Figure \ref{fig:acc_vs_nmi_alldata} shows how NeF representations obtained using the shared initialization already display a semantic structure. For NeFs trained under the shared initialization scheme, when measuring the mutual information between the assignment obtained with simple k-means and the ground truth labels, we can see how semantically similar neural fields are clustered together. 
Overall, a higher NMI between these two assignments results in higher downstream classification accuracy.

\begin{figure}[t]
    \centering
    \includegraphics[width=\linewidth]{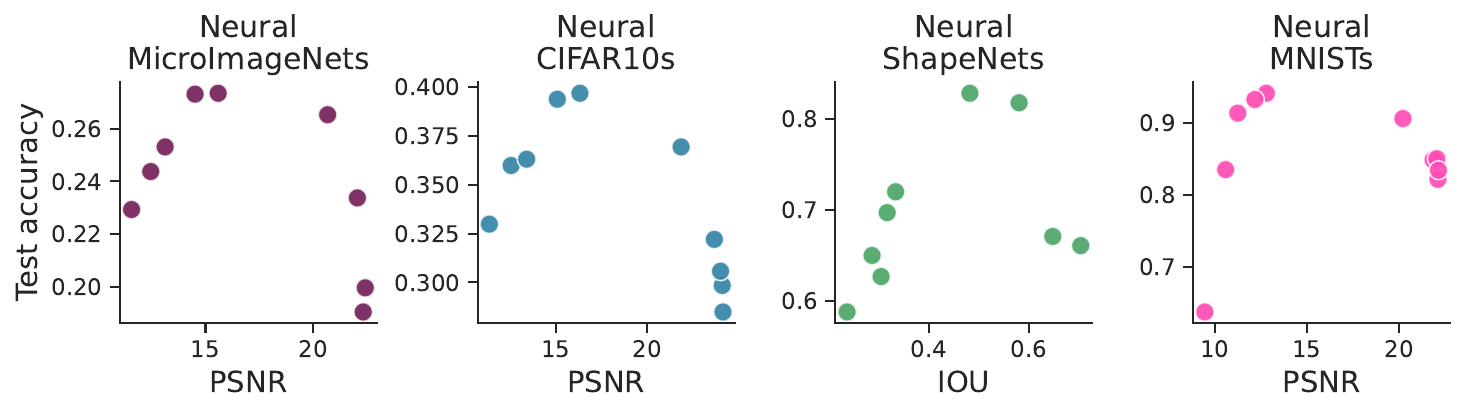}
    \caption{Results for the \textit{reconstruction quality} ($\uparrow$) vs \textit{test accuracy} ($\uparrow$) experiment. By fixing the NeF's architecture we can more clearly see that there is a trade-off between visual quality and classification accuracy.}
    \label{fig:acc_vs_psnr}
\end{figure}
\begin{figure}[t]
    \centering
    \includegraphics[width=\linewidth]{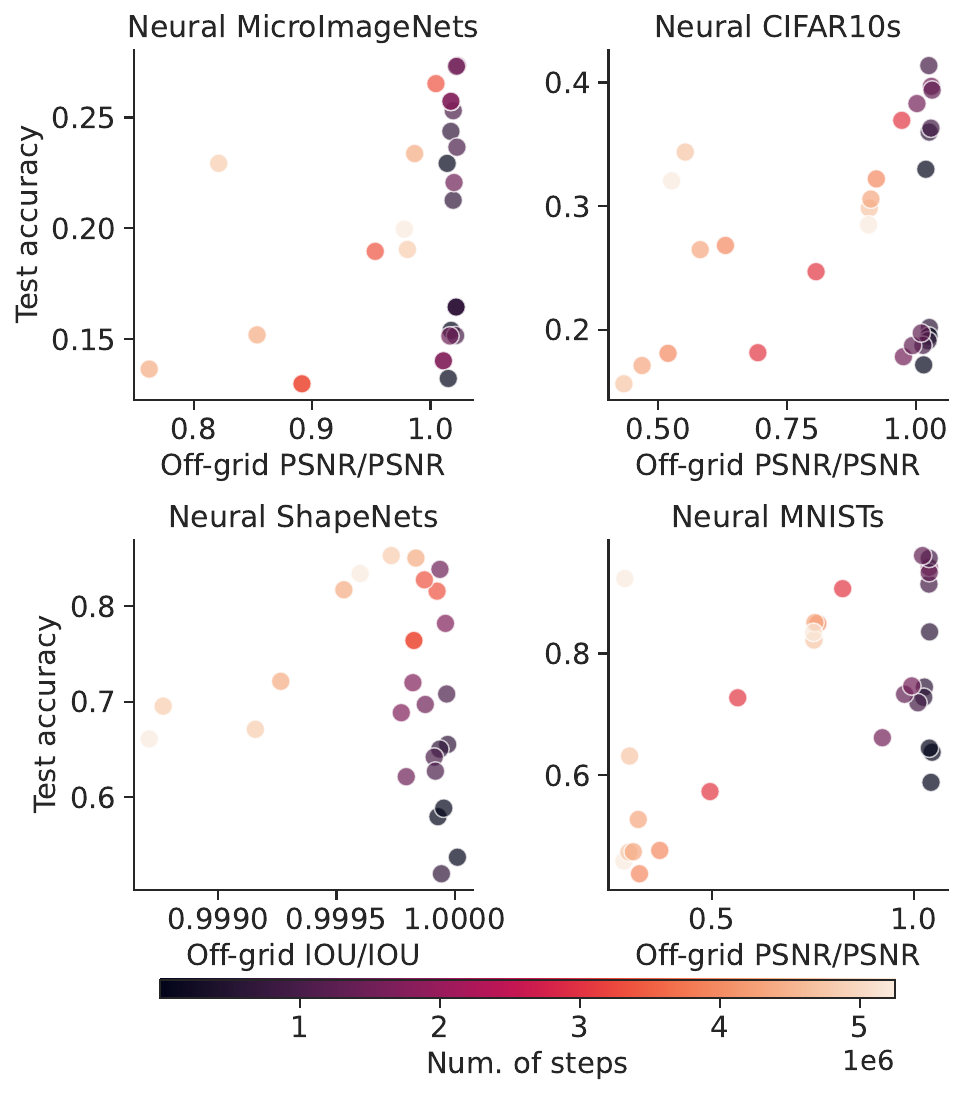}%
    \caption{We fit 220 Neural Datasets using different hidden dimensions and the number of steps while keeping the same shared initialization. We find that the ratio of off-grid reconstruction quality and in-grid reconstruction quality can be used to form a heuristic that correlates with high test accuracy.}
    \label{fig:acc_vs_psnr_ratio_all}
\end{figure}

\subsection{Overtraining Neural Fields}\label{sec:experiment2}

\paragraph{Description.}
We experiment with different hidden dimensions and number of steps with shared initialization to study the relationship between reconstruction quality and representation quality. We observe the relationship between reconstruction quality, off-grid reconstruction quality, and classification accuracy.

\paragraph{Analysis.} 
The PSNR and IOU measure the reconstruction quality of neural fields, whereas the off-grid PSNR and off-grid IOU measure overfitting. We plot the classification accuracy achieved with Neural Datasets of different reconstruction quality in Fig. \ref{fig:acc_vs_psnr}. We observe that the classification accuracy increases with reconstruction quality but then reaches a maximum and starts decreasing. The general trend is consistent across different datasets, architectures, and optimization choices. In Fig. \ref{fig:acc_vs_psnr_ratio_all} we plot the classification accuracy against the ratio $\frac{\texttt{off-grid-PSNR}}{\texttt{on-grid-PSNR}}$ for images or $\frac{\texttt{off-grid-IOU}}{\texttt{on-grid-IOU}}$ for shapes. Interestingly, a clear pattern emerges that is consistent across datasets: when the number of steps is the only hyperparameter that is changed, stopping as soon as the NeF starts showing signs of overtraining leads to the best representation quality. We conclude that overfitting negatively affects downstream performances.

\subsection{Expressivity of Neural Field Architectures}

\paragraph{Description.} We experiment with the same setting of Sec. \ref{sec:experiment2}. This time we focus on the effects of the architecture choice on test accuracy and reconstruction quality. 

\begin{figure}[t]
    \centering
    \includegraphics[width=.9\linewidth]{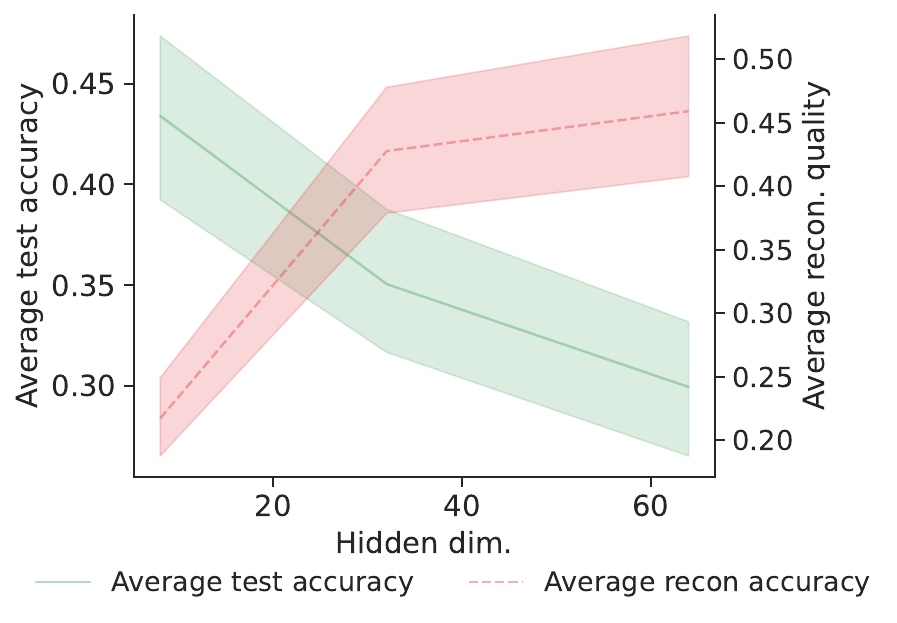}%
    \caption{We measure the average test accuracy across all Neural Datasets and different numbers of steps. Additionally, we normalize the PSNR for the image datasets and the IOU for the shape dataset to measure the average reconstruction quality.}
    \label{fig:acc_vs_hidden_dim}
\end{figure}

\paragraph{Analysis.} Figure \ref{fig:acc_vs_hidden_dim} aggregates results from 110 experiments across the different datasets and shows how increasing the hidden dimension tends to lead to worse representation quality while the higher expressivity of the NeF leads to higher PSNR. When exploring different activation functions and model layouts (see supplementary material) we observe a consistent behavior.

\section{Benchmarks}
\label{sec:benchmarks}

Part of the fast progress of Deep Learning is due to publicly available datasets for benchmarking that allow for fair comparisons. Neural representations are a relatively new approach to representation learning and because of that, we lack good benchmark datasets to compare architectures that perform different downstream tasks on them. Since we observed great variability in downstream performance as we vary the NeF architecture and optimization strategy, proper benchmarking is key to systematic progress hereupon.

After having explored the best optimization and architecture for several datasets, we now establish a new benchmark with existing downstream models that had previously achieved the state-of-the-art classification using neural representations.

Table \ref{tab:sota-our} shows how state-of-the-art results can be achieved with our simple GNN baseline just by choosing the correct neural representation parameters. 

We create Neural Datasets variants for CIFAR10, MNIST, MicroImageNet and ShapeNetvs-10, and bundle these as first entries in the Neural Field Arena. With these Neural Datasets, we evaluate the performance of DWS~\cite{navon2023equivariant}, a Relation Transformer~\cite{zhang2023neural}, and our simple GNN baseline.

\begin{table}
\centering
\resizebox{\columnwidth}{!}{%
\begin{tabular}{ld{2.4}d{2.4}d{2.4}}
\toprule
\textbf{Neural Dataset} & \multicolumn{3}{c}{Test accuracy (\%)} \\
\cmidrule{2-4}
& \mc{GNN baseline} & \mc{DWS \cite{navon2023equivariant}} & \mc{Transformer \cite{zhang2023neural}} \\
\midrule
\textit{Neural MNIST \cite{navon2023equivariant}} & - & 85.70\spm{0.60} & 92.40\spm{0.30} \\
\textit{Neural MNIST} (Ours) & \boldc{96.40 \spm{0.11}} & \boldc{93.51\spm{0.12}} & \boldc{95.08\spm{0.95}} \\
\bottomrule
\end{tabular}
}
\caption{We compare the proposed neural MNIST to previous work~\cite{navon2023equivariant}. The insights gained in the study have resulted in a significant performance increase when the same models are used. We note that no hyper-parameter tuning was performed.}
\label{tab:sota-our}
\end{table}

\begin{table}\label{tab:benchmark}
\centering
\resizebox{\columnwidth}{!}{%
\begin{tabular}{ld{2.4}d{2.4}d{2.4}}
\toprule
\textbf{Neural Dataset} & \multicolumn{3}{c}{Test accuracy (\%)} \\
\cmidrule{2-4}
& \mc{GNN baseline} & \mc{DWS \cite{navon2023equivariant}} & \mc{Transformer \cite{zhang2023neural}} \\
\midrule
\textit{Neural CIFAR10} (Ours) & 39.83 \spm{1.70} & 44.01\spm{0.48} & 44.11\spm{0.20} \\
\textit{Neural MicroImageNet}  (Ours) & 25.08 \spm{0.01} & 28.61\spm{0.48} & 28.50\spm{0.09} \\
\textit{Neural ShapeNet}  (Ours) & 82.96 \spm{0.02} & 91.06\spm{0.25} & 90.31\spm{0.15} \\
\bottomrule
\end{tabular}
}
\caption{We benchmark the new three proposed neural datasets using state-of-the-art methods from literature~\cite{navon2023equivariant, zhang2023neural}.}
\end{table}

\section{Related Work}
\label{sec:related_work}

\paragraph{Neural Fields.} Recent years have seen a surging interest in the development of methods to facilitate NeF-based signal reconstructions. A number of works explore changes in neural network architectures to address the \textit{spectral bias} of conventional ReLU MLPs towards low-frequency signals \cite{rahaman2019spectral}, which impedes the learning of high-fidelity signal reconstructions in fine detail. \citet{tancik2020fourier} use a random Fourier feature mapping applied to the input of a NeF to prevent over-smoothing, whereas \citet{sitzmann2020implicit} propose to use periodic activations to artificially inflate the frequency spectrum of the network, achieving better modeling of fine details. Multiplicative filters networks \citep{fathony2020multiplicative} are a different class of functions that avoids traditional compositional network depth and multiplies together simple functions, which has the added benefit of direct control over the network's frequency spectrum \cite{romero2021flexconv}. Importantly, all of these works only investigate NeF architectures for improved signal reconstruction quality. We instead focus on researching NeF architectures for their viability as \textit{downstream representations}, i.e. learning on the parameter space of NeFs.

A different direction of research attempts to alleviate the computational burden of fitting individual NeFs to dataset samples, and to reduce the complexity of a NeF as representation. Building on the concept of neural network conditioning through modulation \cite{perez2018film}, \citet{chan2021pi, mehta2021modulated} propose to use a single NeF to represent a diversity of signals. The activations of this NeF, which shares parameters across the entire dataset, are modulated with datapoint-specific low-dimensional latents to yield datapoint-specific outputs. The latents may be obtained using an encoder \cite{mehta2021modulated}, or learned by using a combination of meta-learning and auto-decoding \cite{park2019deepsdf}. \citet{dupont2022data} provided the insight that these learned latents may be used as downstream representation instead of the parameters of a sample-specific NeF to simplify the task of learning on a parameter space. Although reducing the complexity of learning on parameter space, this approach limits performance -- both in terms of signal reconstruction quality and downstream performance \cite{bauer2023spatial}. \citet{bauer2023spatial} show that a spatial grid of latent vectors is needed to reach competitive downstream performance, entailing all limitations that come with discrete data representations. In this work --to obtain the most general and flexible signal parameterizations-- we instead focused on NeFs without shared parameters or sample-specific conditioning and explored what factors facilitate learning on their parameter space directly in downstream tasks.

\paragraph{Applications of neural representations.} We briefly mention a few applications of neural representations. Several works propose their use for generating shapes using diffusion models \cite{erkocc2023hyperdiffusion}. In the context of dynamical systems and PDEs, \cite{yin2022continuous} show that neural representations can be used for dynamics forecasting to the flow of partial differential equations as spatially and time-continuous functions.

\paragraph{Deep Weight Space methods.}
The exploration of deep learning architectures for processing neural network parameters is relatively recent. Here, we offer an overview of the key studies. The works by \cite{eilertsen2020classifying} and \cite{unterthiner2020predicting} revolve around predicting characteristics of trained neural networks (NNs) by scrutinizing their weight configurations. \cite{eilertsen2020classifying} is primarily focused on estimating the hyperparameters used during network training, while \cite{unterthiner2020predicting} is dedicated to evaluating a network's generalization capabilities. Both investigations involve applying conventional neural networks to either the flattened weight data or their statistical attributes. On the other hand, \cite{xu2022signal} introduced a novel concept where NNs are processed using another neural network that operates on a combination of their high-order spatial derivatives. This technique is particularly suitable for NeF representations where derivative information plays a crucial role. Nevertheless, the adaptability of these networks to broader tasks remains uncertain, and the demand for high-order derivatives can introduce significant computational overhead. Lastly, the works of \cite{zhang2023neural}, proposed a transformer-like architecture for compatibility with multiplicative filters networks, and the studies by \cite{navon2023equivariant} and \cite{zhou2023permutation}, which initially emphasized the importance of augmentations and permutation-invariant architectures for processing neural field weights.

\section{Conclusion}
The goal of this work is to gain a deeper understanding of NeFs as data representations. To this end, we developed a library for fast generation and manipulation of Neural Datasets for different modalities, such as images and shapes. The speed gain in fitting Neural Datasets is substantial compared to naive approaches, practically enabling the use of neural fields for representing datasets at scale.

The study we conducted resulted in significant insights into what constituted better-quality NeF representations.
The first key takeaway is that obtaining good NeF representations is a trade-off between good reconstruction and how spread NeFs are in parameter space. Generally, to achieve good reconstruction more gradient descent iterations are needed, which would spread the NeF representations in parameter space, negatively affecting their quality. Because of that, we propose to fit Neural Datasets from the same \textit{shared initialization}, and demonstrate that this simple practice results in significantly higher representation quality across different modalities.
Second, \textit{overfitting hurts representation quality}. In other words, good reconstruction helps downstream performances up until the NeFs are overfitting the signals. To address this, we propose to stop fitting the representations early, before overfitting happens.
Finally, we show how an increase in the number of parameters used to fit a signal leads to a decrease in downstream performance.\looseness=-1

Last, we tackle the lack of benchmarks to compare architectures that perform different downstream tasks on NeF representations by providing the \textit{Neural Field Arena}, covering various modalities.

\vspace{2mm}\noindent \textbf{Limitations.}
We limited our study to images and shapes leaving out neural radiance fields (NeRFs) and videos despite our library being suited to those types of NeFs.
Additionally, for practical reasons, we choose to measure representation quality of NeFs using the classification accuracy of a GNN classifier. Although classification is often used in the representation learning literature to evaluate representation quality~\cite{bengio2013representation, chen2020simple,locatello2019challenging}, we acknowledge that good classification accuracy might not result in good downstream performance for general tasks. We plan to explore other downstream tasks and include them in the Neural Field Arena in future work.
{
    \small
    \bibliographystyle{ieeenat_fullname}
    \vspace{-2\baselineskip}\bibliography{main}
}

\clearpage
\setcounter{page}{1}
\setcounter{section}{0}
\maketitlesupplementary

\section{NeFs Architectures}

\paragraph{SIREN \cite{sitzmann2020implicit}.} SIRENs, short for sinusoidal representation networks, are a variant of multilayer perceptron characterized by the utilization of sinusoidal activations. By employing periodic activation functions, SIRENs enhance the network's capability to grasp nuances in the signal, particularly when handling wave-based signals and images. The configuration of a SIREN network is the following:
\begin{equation}
\scalebox{0.8}{
$\begin{aligned}
    &h^{(1)} = x \\
    &h^{(i)} = \sin\left(\Omega_0  W^{(i-1)}h^{(i-1)} + b^{(i-1)}\right), \, i=2, \dots, k-1 \\
    &f_{\theta}(x) = W^{(k)}h^{(k)} + b^{(k)},
\end{aligned}$}
\end{equation}
where $W^{(i-1)} \in \mathbb{R}^{d_{i} \times d_{i-1}}$ and $b^{(i-1)} \in \mathbb{R}^{d_i}$ denote the weight matrix and bias vector for the $i$-th layer, respectively, and $\Omega_o$ is a scalar. Here, the $\sin$ function is utilized as the activation function at each layer.

\paragraph{MFNs \cite{fathony2020multiplicative}.} \emph{Multiplicative Filters Networks} are NeFs architectures that don't rely on compositional depth for expressivity, but rather on nonlinear filters that are applied to the input and iteratively passed through linear functions and multiplied together. In particular, an MFN is defined by the recursion
\begin{equation}
\label{eq:RecursiveFormulaMFN}
\scalebox{0.8}{
$\begin{aligned}
    &z^{(1)} = g\left(x; \psi^{(1)}\right) \\
    &z^{(i+1)} = \left(W^{(i)}z^{(i)} + b^{(i)} \right) \odot g\left(x; \psi^{(i+1)}\right), \, i=1, \dots, k-2 \\
    &f_\theta(x) = W^{(k-1)}z^{(k-1)} + b^{(k-1)},
\end{aligned}$}
\end{equation}
where $\odot$ is the element-wise multiplication, $W^{(i)} \in \mathbb{R}^{d_{i+1} \times d_i}$, $b^{(i)} \in \mathbb{R}^{d_{i+1}}$ and $g : \mathbb{R}^{d} \rightarrow \mathbb{R}^{d_i}$ are the nonlinear filters parameterized by $\psi_i$ that are applied to the input.

In this paper we use a linear layer composed with a sine function as filters: $g\left(x; \psi^{(i+1)}\right) = \sin{\omega x + \phi}$ where $\omega_i \in \mathbb{R}^{d\times d_1}$ and $\phi_i \in \mathbb{R}^{d_1}$. Interestingly, one can show that such a multiplicative filter network is equivalent to a linear function of an exponential (in $k$) number of Fourier basis functions.

\paragraph{RFFNets \cite{tancik2020fourier}.} \emph{Random Fourier Features Networks} leverage the Random Fourier Features technique, initially introduced in machine learning, to approximate kernel methods efficiently. In the context of neural fields, RFFNet uses random Fourier features to approximate the feature maps resulting from the kernel functions. Instead of explicitly computing the kernel function, which can be computationally intensive for high-dimensional data, RFFNet employs random projections to map input data into a higher-dimensional space. These random projections mimic the feature space resulting from a kernel function, such as the Gaussian or radial basis function (RBF) kernel. The expression of a RFFNet is
\begin{equation}
\scalebox{0.8}{
$\begin{aligned}
    &h^{(1)} = \sqrt{\sigma} W^{(0)}x \\
    &h^{(2)} = [\sin(z^{(1)}), \, cos(z^{(1)})] \\
    &h^{(i)} = \text{ReLU}\left(W^{(i-1)}h^{(i-1)} + b^{(i-1)}\right), \, i=3, \dots, k-1 \\
    &f_{\theta}(x) = W^{(k)}h^{(k)} + b^{(k)},
\end{aligned}$}
\end{equation}
where $W^{(0)} \in \mathbb{R}^{d_1 \times 2}$ and $\sqrt{\sigma}$ is also called the \texttt{std} and is a hyperparameter that controls the frequency range of the embedding layer.
\paragraph{Metrics for reconstruction quality.}
\emph{Peak Signal-to-Noise Ratio (PSNR)} is a widely used metric in signal processing that measures the quality of a reconstructed or processed signal concerning the original signal. It evaluates the fidelity of the reconstructed signal by comparing the maximum possible power of the original signal to the power of the difference between the original and reconstructed signals, expressed in decibels (dB). A higher PSNR value indicates a smaller difference between the signals, suggesting better reconstruction quality and less distortion. 

The \emph{Intersection over Union, (IOU)} is a metric often used in shape reconstruction or object detection tasks within computer vision. In the context of shape reconstruction, the IOU measures the overlap between the predicted shape (such as a bounding box, mask, or region) and the ground truth shape in an image. It calculates the ratio between the area of overlap and the area of union between the predicted and ground truth shapes. Higher IOU values, closer to 1, indicate a better match between the predicted and actual shapes, signifying more accurate reconstruction or detection.
\section{NeFs Initialization}
\paragraph{SIRENs.} As pointed out by the authors, if not initialized properly, a SIREN results in poor reconstruction. We follow the principled initialization scheme proposed by the authors aimed at maintaining the activation distribution across the network, ensuring that the initial output remains independent of the number of layers. This can be done by sampling the rows $w_i$ of the weight matrices from a uniform distribution: 
\begin{equation}
    w_i \sim \mathcal{U} \left(-\cfrac{c}{\sqrt{\text{fan\_in}}}, \cfrac{c}{\sqrt{\text{fan\_in}}}\right).
\end{equation}
We refer to \cite{sitzmann2020implicit} for the derivations. 

\paragraph{MFNs.} The FourierNet has similar behavior to the SIREN network, as $\sin$ activation functions are being used as filters. The linear layers -- i.e. $W^{(i)}, b^{(i)}, i=1,\dots , k-1$ -- are initialized with the same scheme as a SIREN. However, the number of filters will now directly affect the final result, as these are not acting sequentially upon each other, instead, they are a multiplicative factor that affects each linear layer element-wise. Therefore, the filters are initialized according to the following:
\begin{equation}
    w_i \sim \mathcal{U} \left(-\sqrt{\cfrac{\text{s}}{k-1}}, \sqrt{\cfrac{s}{k-1}}\right),
\end{equation}
where $k-1$ is the number of filters and $s$ is an input scaling constant. The latter is a hyperparameter that has a similar effect to the \texttt{std} of the RFFNet, which we fix to $16$ in our experiments.

\paragraph{RFFNets.} For the RFFNet, we choose to initialize the coefficients of the embedding using a standard Gaussian, such that $ W^{(0)} \sim \mathcal{N}(0, I)$. Then, the layers are initialized using the uniform version of Kaiming He's initialization:

\begin{equation}
    w_i \sim \mathcal{U} \left( -\sqrt{\cfrac{6}{\text{fan\_in}}}, \sqrt{\cfrac{6}{\text{fan\_in}}} \right),
\end{equation}

which is a uniform distribution with variance of $2/\text{fan\_in}$.
\begin{table}[t]
\centering
\resizebox{\columnwidth}{!}{%
\begin{tabular}{@{}lcl@{}}
\toprule
Hyperparameter  & Range & Range type  \\
\midrule
Initialization  & \texttt{[True, False]} & Categorical \\
Weight decay    & \texttt{[1e-8, 1e-2]} & Logarithmic \\
Learning rate   & \texttt{[1e-5, 1e-1]} & Logarithmic \\
Number of steps & \texttt{[20, 200, 1000, 5000, 10000, 20000]} & Categorical\\
\bottomrule
\end{tabular}
}
\caption{Range of the \textit{optimization} hyperparameters used for the initial study.}
\label{tab:initial_hyper_optim}
\end{table}

\begin{table}[t]
\centering
\resizebox{\columnwidth}{!}{%
\begin{tabular}{@{}llcl@{}}
\toprule
NeF& Hyperparameter  & Range & Range type  \\
\midrule
\multirow{3}{*}{SIREN}  & Hidden dim. & \texttt{[8, 128]} & Logarithmic \\
& Num. layers     & \texttt{[3, 6]} & Linear      \\
& $\Omega_0$      & \texttt{[1, 1e2]} & Logarithmic \\
\midrule
\multirow{3}{*}{RFFNet} & Hidden dim. & \texttt{[4,128]} & Logarithmic \\
& Num. layers     & \texttt{[3, 6]} & Linear      \\
& $\sqrt{\sigma}$ & \texttt{[1e-2, 1e2]} & Logarithmic \\
\midrule
\multirow{2}{*}{MFN} & Hidden dim. & \texttt{[8, 128]}    & Logarithmic \\
& Num. filters & \texttt{[1, 6]} & Linear  \\
\bottomrule
\end{tabular}
}
\caption{Range of the \textit{architecture} hyperparameters used for the initial study.}
\label{tab:initial_hyper_arch}
\end{table}

\begin{table}[t]
\centering
\resizebox{\columnwidth}{!}{%
\begin{tabular}{@{}llc@{}}
\toprule
NeF& Hyperparameter  & Range \\
\midrule
\multirow{6}{*}{SIREN}  & Hidden dim. & \texttt{[8, 32, 64]}  \\
& Num. steps     & \texttt{[5, 15, 25, 50, 75, 1000, 5000, 10000, 20000, 50000]}       \\
& Learning rate & \texttt{1e-3} \\
& Weight decay & \texttt{0} \\
& Num. layers & \texttt{3} \\
& $\Omega_0$      & \texttt{9} \\
\midrule
\multirow{6}{*}{RFFNet} & Hidden dim. & \texttt{[16, 32, 64]} \\
& Num. layers     & \texttt{[5, 15, 25, 50, 75, 1000, 5000, 10000, 20000, 50000]}      \\
& Learning rate & \texttt{1e-4} \\
& Weight decay & \texttt{0} \\
& Num. layers & \texttt{5} \\
& $\sqrt{\sigma}$ & \texttt{1e-1}  \\
\midrule
\multirow{5}{*}{MFN} & Hidden dim. & \texttt{[16, 32, 64]}     \\
& Num. filters & \texttt{[5, 15, 25, 50, 75, 1000, 5000, 10000, 20000, 50000]}  \\
& Learning rate & \texttt{5e-3} \\
& Weight decay & \texttt{0} \\
& Num. filters & \texttt{4} \\
\bottomrule
\end{tabular}
}
\caption{Parameters used during the second phase of the study. These are chosen based on the results obtained during the first phase.}
\label{tab:second_phase_params}
\end{table}
\section{The Classifier Architectures}
Our classifier network is inspired by the work of \cite{zhang2023neural}, where the authors propose to use the computational graph of neural networks and encode NeF representations with graph networks or transformers that are invariant to the permutation symmetries present in the parameter space.
We construct a message-passing GNN where the biases of each NeF layer correspond to node features, while the weights correspond to edge features.
% For a standard fully-connected MLP, the graph edge features matrix is organized as a block-superdiagonal matrix, i.e. a block matrix with blocks populated 1 above and to the right of the main diagonal (see Figure \ref{fig:edge_feature_matrix}).
The authors also extend transformer architectures, in particular PNA \citep{corso2020principal}, and Relational Transformer \citep{diao2023relational}. In the benchmarks, the Transformer mentioned is a Relational Transformer.
The DWSNet is again a permutation invariant architecture that has been proposed in \cite{navon2023equivariant}. It is a composition of several different linear equivariant layers, such as DeepSets \cite{zaheer2017deep}, and pointwise nonlinearities. To obtain an invariant classifier we simply compose an equivariant DWSNet with an invariant linear layer, eventually follwd by a fully connected MLP for expressivity.

The GNN uses $4$ steps of message passing, with a hidden dimension of $64$. The MLP used for the update function on the nodes has $3$ layers and $256$ hidden dimension, while the one used for the edges has $2$ layers and $256$ hidden dimension.

\begin{figure}
    \centering
    \includegraphics[width=\columnwidth]{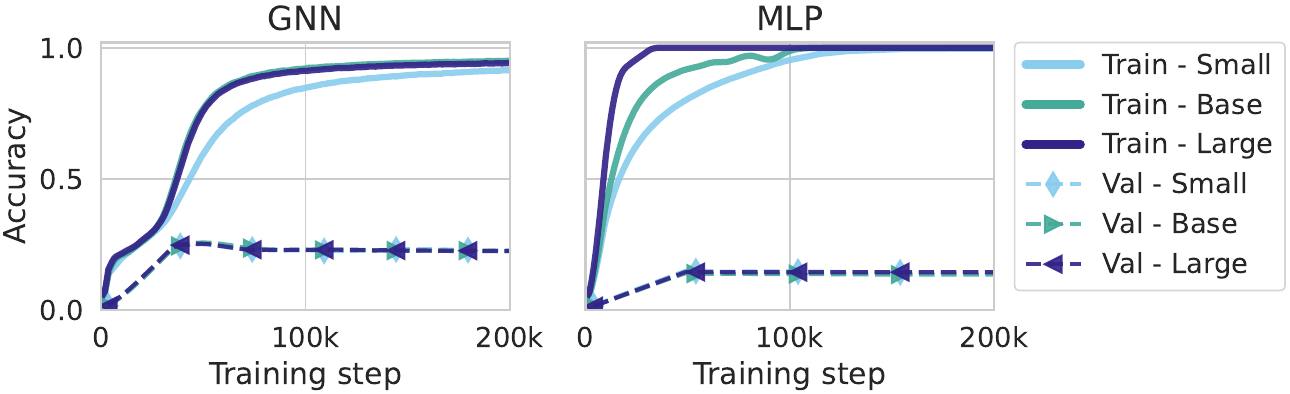}
    \caption{Training and test accuracy for three different-size GNN and MLP models on a CIFAR10 neural dataset. Overfitting ability is clear also for small capacities.}
    \label{fig:accuracies}
\end{figure}

\subsection{Ablation on the Architectures}

To evaluate the ability of the models to overfit to the neural representations, we carried out an ablation study using both an MLP and a GNN. The MLP uses linear layers with ReLU activations and batch normalization~\cite{ioffe2015batch}. The GNN is the one previously described. For both, we decide to train a small/\textit{base}/large versions. For the MLP these correspond to 74k/116k/363k parameters, while for the GNN they correspond to 500k/1.2M/1.5M parameters. In both cases, we use a CIFAR10 dataset from our study and simply use the neural representations as inputs. For the MLP the input is the concatenated and flattened version of the weights and biases, while for the GNN it is the actual graph of the neural field, where the connectivity is determined by the weights, and node features by the biases. We show the training and validation curves in Fig. \ref{fig:accuracies}. These show clear signs of overfitting, the training curve of the GNN slowly approaches the maximum accuracy, while the one of the MLP does so quickly. For both, the validation curves are almost identical at all scales, likely due to capacity saturation.

\section{Extent of the Study}

In total, the study we carried out required performing around 30k experiments, each corresponding to a new neural dataset being fit, and a classifier being trained on it.
The study was split into two phases. 

In the initial phase, we searched a large space of hyperparameters using a smaller subset of the whole dataset, and trained the classifier only for $10$ epochs. In the initial hyperparameter exploration, we looked at \textit{optimization} -- see Tab.~\ref{tab:initial_hyper_optim} -- and \textit{architecture} hyperparameters -- see Tab.~\ref{tab:initial_hyper_arch}. For this exploration, we used the TPE sampler implemented in Optuna~\cite{bergstra2011algorithms,akiba2019optuna}. In short, this sampler uses the previous results to inform the choice of the best hyperparameters to use next. Each experiment took between a few minutes to a few hours to run, depending on the number of steps and the size of the architecture. This was done for all $4$ datasets and $3$ models. 

In the second phase, we ran a full grid search over the initialization scheme, the number of steps, and the hidden dimension of the NeFs. These were the most impactful parameters according to the search performed in the first phase. This grid search was performed for all datasets and models. Refer to Tab.~\ref{tab:second_phase_params}.
\onecolumn
\section{Additional Experiments}

Follows additional experiments carried out on FourierNet (MFN) and RFFNet. The results align with what was found when using SIREN, which suggests that these findings hold true across architectures. It is important to remember that, although the architectures tested are different and employ different activation functions, they are still very similar to each other, as the fundamental building block is fully connected feed-forward neural networks, with a small number of layers. 

\begin{figure*}[h]
    \centering
    \begin{subfigure}[b]{0.8\textwidth}
        \centering
        \includegraphics[width=\textwidth]{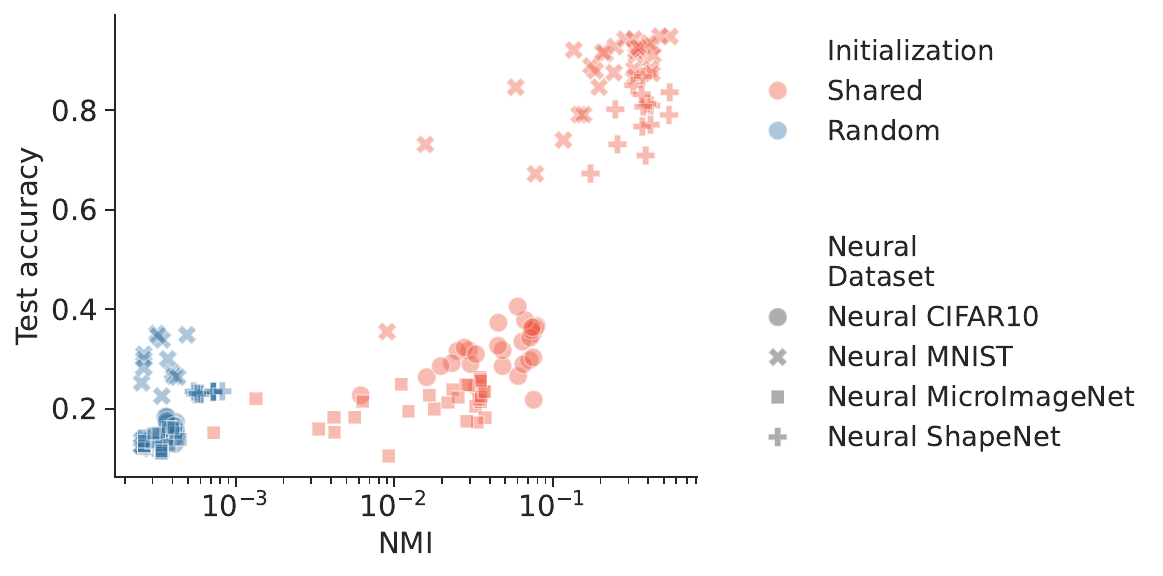}
        \caption{RFFNet}
    \end{subfigure}
    \begin{subfigure}[b]{0.8\textwidth}
        \centering
        \includegraphics[width=\textwidth]{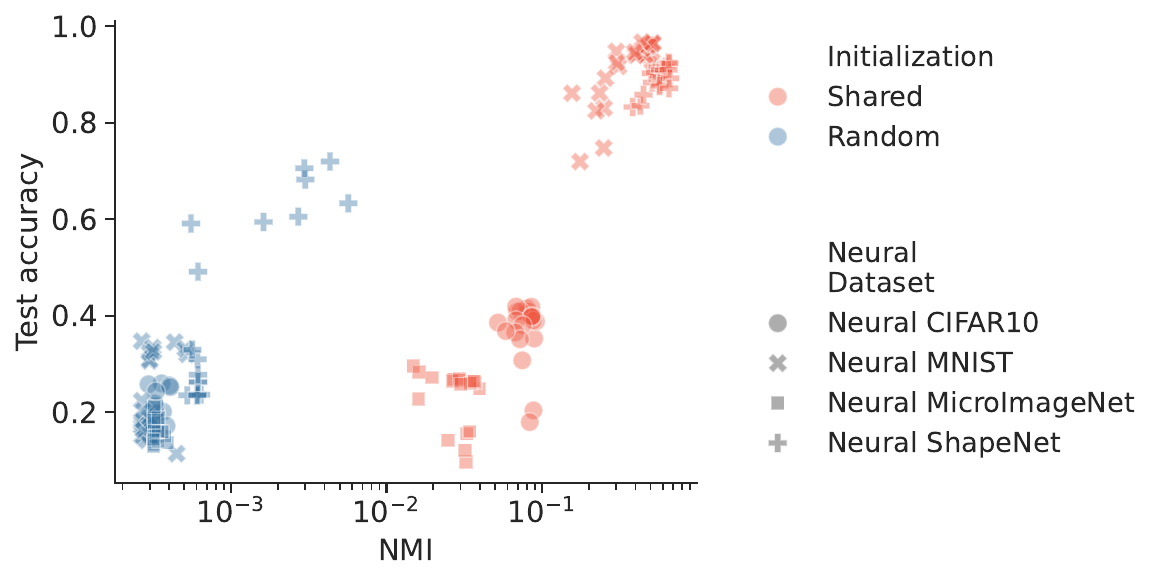}
        \caption{MFN}
    \end{subfigure}
    \caption{Results of the \textit{test accuracy} ($\uparrow$) vs \textit{NMI} ($\uparrow$) using different initialization on 220 Neural Datasets created using different hidden dimensions and the number of steps with RFFNets and MFNs. Different datasets are stylized using different markers. Shared initialization leads to semantically structured NeF representation and, generally to better performance. Both RFFNet and MFN achieve the same separation when using the shared initialization, as SIREN did.}
    \label{fig:rffnet_mfn_acc_vs_nmi_alldata}
\end{figure*}
\begin{table*}[h]
\centering
\resizebox{.7\linewidth}{!}{%
\begin{tabular}{@{}llld{2.4}d{3.2}@{}}
\toprule
NeF & Neural dataset & Initialization & \mc{Test accuracy} & \mc{Gain (\%)} \\
\midrule
\multirow{2}{*}{RFFNet} & \multirow{2}{*}{\textit{Neural MNISTs}} & Random & 0.20\spm{0.09} & \\
  &  & Shared & \boldc{0.86\spm{0.12}} & 327.10\\
\midrule
\multirow{2}{*}{RFFNet}  & \multirow{2}{*}{\textit{Neural CIFAR10s}} & Random & 0.16\spm{0.02} & \\ 
  &  & Shared & \boldc{0.31\spm{0.05}} & 102.33 \\ 
\midrule
\multirow{2}{*}{RFFNet}  & \multirow{2}{*}{\textit{Neural MicroImageNets}} & Random & 0.13\spm{0.02} & \\
   & & Shared & \boldc{0.21\spm{0.04}} & 60.65\\
\midrule
\multirow{2}{*}{RFFNet}  & \multirow{2}{*}{\textit{Neural ShapeNets}} & Random & 0.23\spm{0.003} & \\
   & & Shared & \boldc{0.79\spm{0.06}} & 239.21 \\
\midrule
% MFN
\multirow{2}{*}{MFN} & \multirow{2}{*}{\textit{Neural MNISTs}} & Random & 0.23\spm{0.08} & \\
  &  & Shared & \boldc{0.91\spm{0.07}} & 304.58\\
\midrule
\multirow{2}{*}{MFN}  & \multirow{2}{*}{\textit{Neural CIFAR10s}} & Random & 0.20\spm{0.03} & \\ 
  &  & Shared & \boldc{0.37\spm{0.06}} & 82.87 \\ 
\midrule
\multirow{2}{*}{MFN}  & \multirow{2}{*}{\textit{Neural MicroImageNets}} & Random & 0.16\spm{0.02} & \\
   & & Shared & \boldc{0.23\spm{0.06}} & 46.39\\
\midrule
\multirow{2}{*}{MFN}  & \multirow{2}{*}{\textit{Neural ShapeNets}} & Random & 0.42\spm{0.19} & \\
   & & Shared & \boldc{0.89\spm{0.03}} & 109.83 \\
\bottomrule
\end{tabular}
}
\caption{Average \textit{test accuracy} (mean $\pm$ standard deviation) and percentage gain across $30$ neural datasets trained using either shared initialization or random initialization with different number of steps and hidden dimensions. The gains achieved with shared initialization are consistent across all tested settings.}
\end{table*}

% \subsection{Shared vs. Random Initialization}

\begin{figure*}[h]
    \centering
    \begin{subfigure}[b]{0.8\textwidth}
        \centering
        \includegraphics[width=\textwidth]{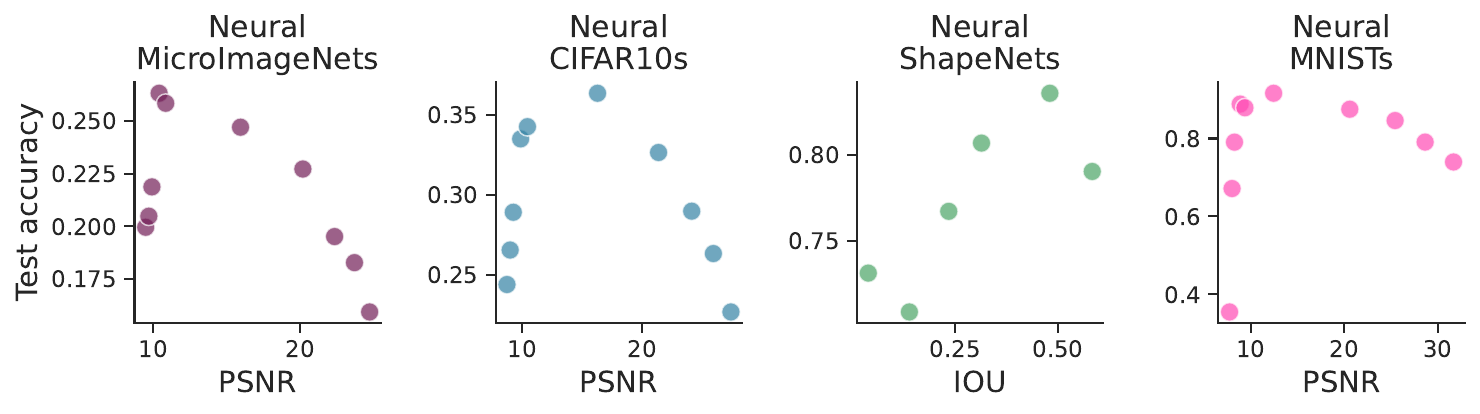}
        \caption{RFFNet}
    \end{subfigure}
    \begin{subfigure}[b]{0.8\textwidth}
        \centering
        \includegraphics[width=\textwidth]{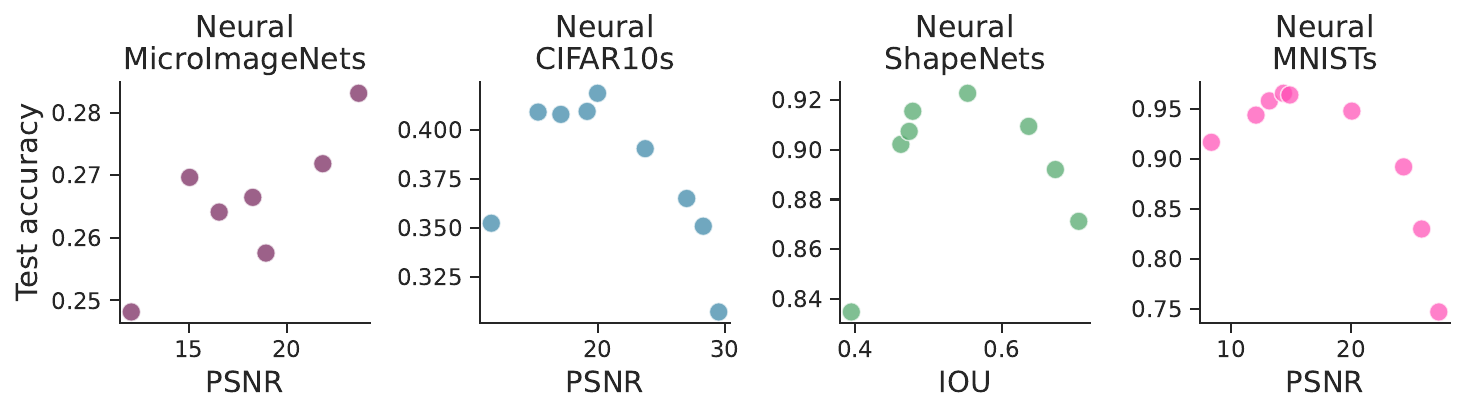}
        \caption{MFN}
    \end{subfigure}
    \caption{Results for the \textit{reconstruction quality} ($\uparrow$) vs \textit{test accuracy} ($\uparrow$) experiment. Similarly to SIRENs, for RFFNets and MFNs, by fixing the NeF's architecture we can more clearly see that there is a trade-off between visual quality and classification accuracy.}
    \label{fig:rffnet_mfn_test_acc_vs_psnr}
\end{figure*}

\begin{figure*}[h]
    \centering
    \begin{subfigure}[b]{0.49\textwidth}
        \centering
        \includegraphics[width=\textwidth]{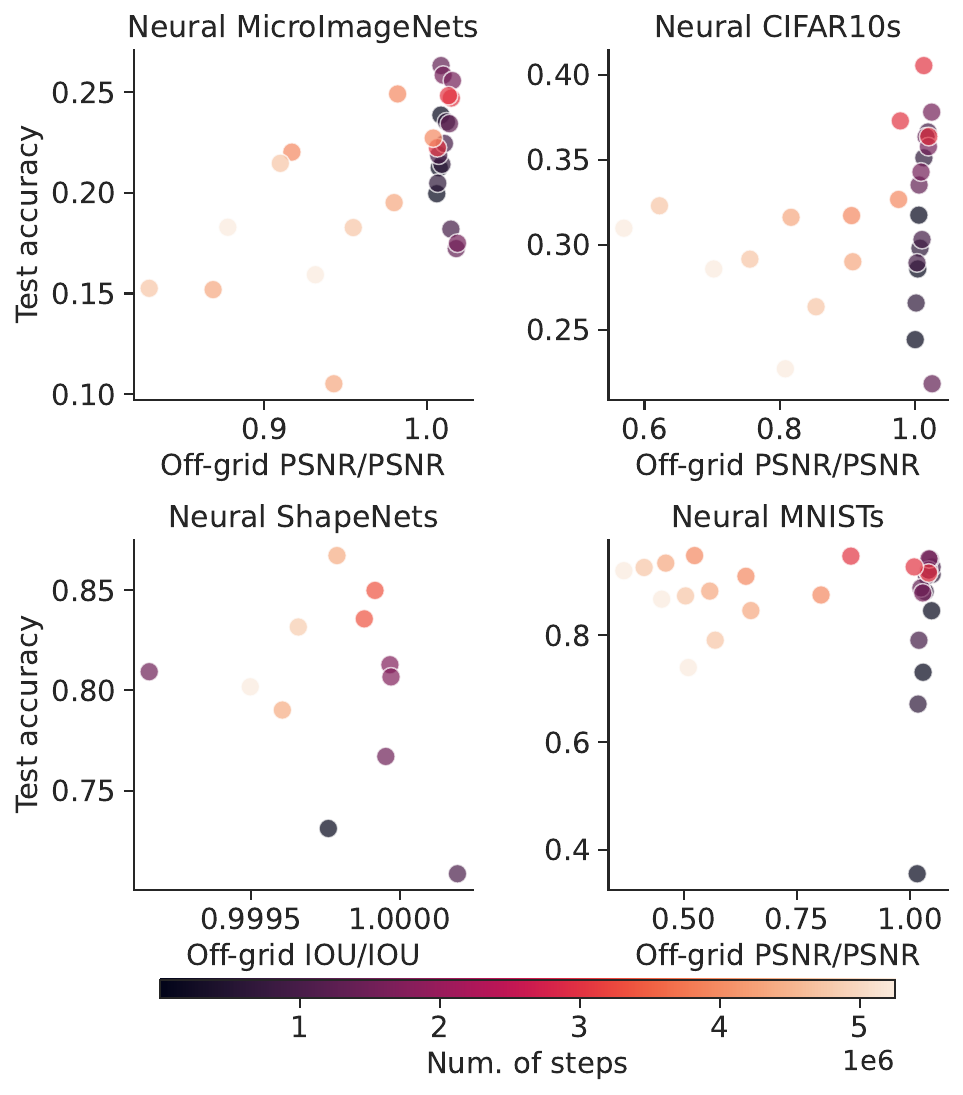}
        \caption{RFFNet}
    \end{subfigure}
    \begin{subfigure}[b]{0.49\textwidth}
        \centering
        \includegraphics[width=\textwidth]{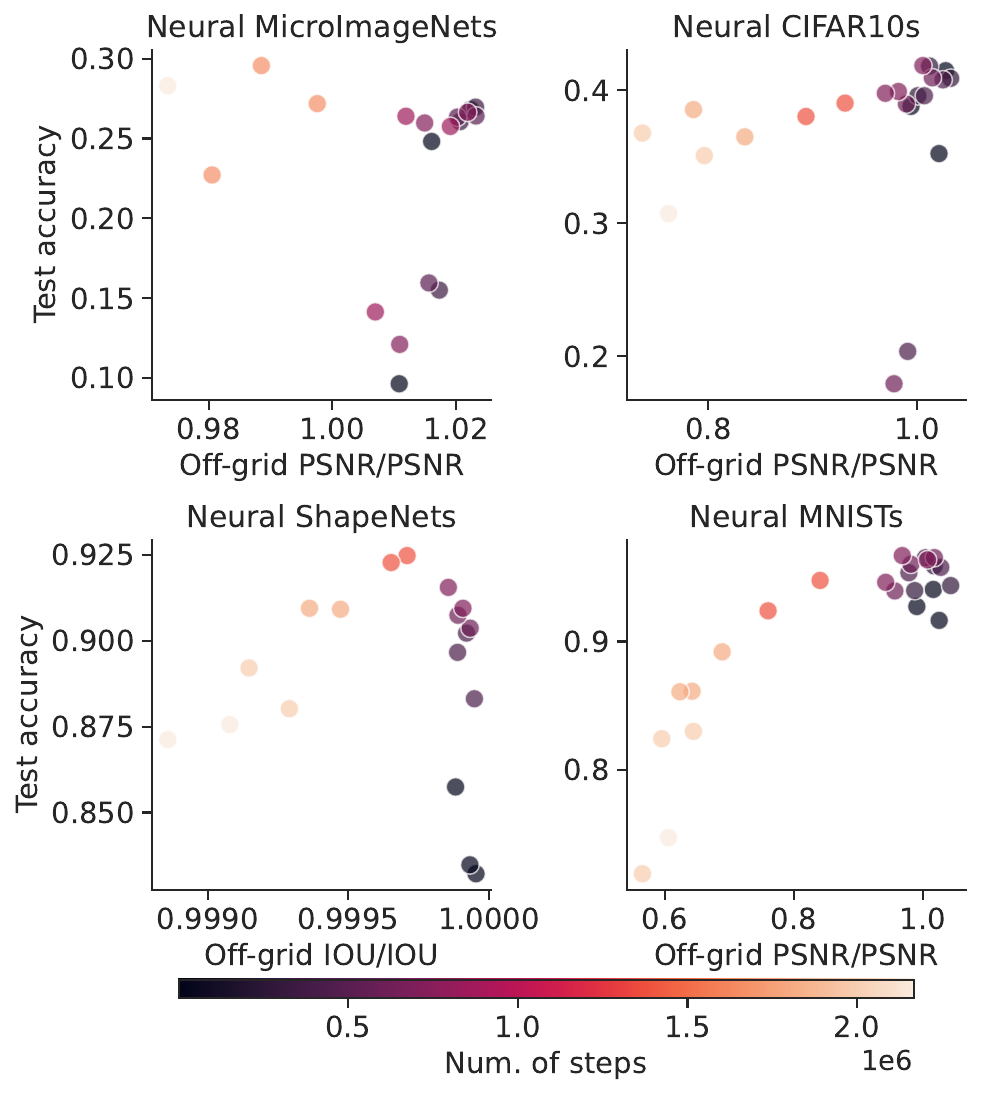}
        \caption{MFN}
    \end{subfigure}
    \caption{We fit 220 neural datasets using different hidden dimensions and number of steps while keeping the same shared initialization. Similarly to SIRENs, for MFNs and RFFNets the ratio of off-grid reconstruction quality and in-grid reconstruction quality can be used to form a heuristic that correlates with high test accuracy. The results across different architectures may be more subtle because of the tuning of different parameters that were not explored here, such as the \texttt{std} in the RFFNet or the input scaling $s$ for the MFN.}
    \label{fig:rffnet_mfn_acc_vs_psnr_ratio_all}
\end{figure*}

\end{document}